\pdfoutput=1
\documentclass[letterpaper, 10 pt, journal, twoside]{IEEEtran}

\IEEEoverridecommandlockouts                            
\usepackage[utf8]{inputenc}
\usepackage{graphicx}
\usepackage{amsmath, amsfonts, amssymb}
\usepackage{algorithm, algpseudocode}
\usepackage{mathrsfs}
\usepackage{rotating}
\usepackage{xcolor}
\usepackage{csquotes}
\usepackage[inline]{enumitem}
\usepackage{booktabs}
\usepackage{multirow}
\usepackage{dblfloatfix}
\usepackage{url}

\newcommand{\change}[1]{{#1}} 							

\newcommand*{\fnref}[1]{\textsuperscript{\ref{#1}}} 	
\newcommand*\rot{\rotatebox[origin=c]{90}}

\definecolor{mydarkblue}{rgb}{0,0.08,0.45}
\definecolor{PorscheRed}{RGB}{213.0, 0.0, 28.0} 
\definecolor{blueP}{RGB}{0.0, 51.0, 102.0}
\definecolor{PorscheGreen}{RGB}{210.0, 235.0, 45.0} 
\definecolor{Green2}{RGB}{210.0, 235.0, 45.0} 
\definecolor{blackP}{RGB}{0.0, 0.0, 0.0} 
\definecolor{grey1P}{RGB}{218.0, 217.0, 222.0}
\definecolor{grey2P}{RGB}{181.0, 180.0, 186.0}
\definecolor{grey3P}{RGB}{115.0, 114.0, 120.0}
\definecolor{grey4P}{RGB}{64.0, 63.0, 69.0}
\definecolor{PorscheYellow}{RGB}{255.0, 192.0, 0.0}

\author{Stefan Löckel$^{1}$, Jan Peters$^{2}$, and Peter van Vliet$^{3}$%
\thanks{Manuscript received September 10, 2019; Revised December 02, 2019; Accepted January 12, 2020.}
\thanks{This paper was recommended for publication by Editor D. Lee upon evaluation of the Associate Editor and Reviewers' comments. This work was supported by Dr. Ing. h.c. F. Porsche AG.}
\thanks{$^{1}$S. Löckel is with the Intelligent Autonomous Systems Group, Technische Universität Darmstadt, 64289 Darmstadt, Germany,
		and also with the Motorsport Department of Dr. Ing. h.c. F. Porsche AG, 71287 Weissach, Germany
        {\tt\footnotesize  stefan.loeckel2@porsche.de}}%
\thanks{$^{2}$J. Peters is with the Intelligent Autonomous Systems Group, Technische Universität Darmstadt, 64289 Darmstadt, Germany,
		and also with the Max-Planck Institute for Intelligent Systems, 72076 Tübingen, Germany
        {\tt\footnotesize  peters@ias.tu-darmstadt.de}}%
\thanks{$^{3}$P. van Vliet is with the Motorsport Department of Dr. Ing. h.c. F. Porsche AG, 71287 Weissach, Germany
        {\tt\footnotesize  peter.van\_vliet@porsche.de}}%
\thanks{Digital Object Identifier (DOI): see top of this page.} } 

\title{A Probabilistic Framework for Imitating Human Race Driver Behavior}

\IEEEtriggercmd{\enlargethispage{-16.3mm}}
\IEEEtriggeratref{10}

\usepackage[absolute,showboxes]{textpos}

\TPMargin{2.5pt}

\newcommand{\copyrightstatement}{
    \begin{textblock}{0.84}(0.08,0.93)    
         \noindent
         \scriptsize
         \copyright  2020 IEEE. Personal use of this material is permitted. Permission from IEEE must be obtained for all other uses, in any current or future media, including reprinting/republishing this material for advertising or promotional purposes, creating new collective works, for resale or redistribution to servers or lists, or reuse of any copyrighted component of this work in other works.
    \end{textblock}
}
\begin{document}
	\maketitle
    \copyrightstatement 
	\begin{abstract}
	Understanding and modeling human driver behavior is crucial for advanced vehicle development. 
	However, unique driving styles, inconsistent behavior, and complex decision processes render it a challenging task, and existing approaches often lack variability or robustness.
	To approach this problem, we propose \textit{Probabilistic Modeling of Driver behavior} (\textit{ProMoD}), a modular framework which splits the task of driver behavior modeling into multiple modules.
	A global target trajectory distribution is learned with Probabilistic Movement Primitives, clothoids are utilized for local path generation, and the corresponding choice of actions is performed by a neural network.
	Experiments in a simulated car racing setting show considerable advantages in imitation accuracy and robustness compared to other imitation learning algorithms.
	The modular architecture of the proposed framework facilitates straightforward extensibility in driving line adaptation and sequencing of multiple movement primitives for future research.
\end{abstract}	
\begin{IEEEkeywords}
	Learning from demonstration, autonomous vehicle navigation	
\end{IEEEkeywords}
	\section{Introduction}
	\IEEEPARstart{R}{eliable} simulations are crucial for modern car development, allowing faster prototyping and a better cost efficiency for both, the design of single parts, and the testing of the overall vehicle performance.
	While vehicle dynamics have been studied and modeled extensively for decades and are well understood even in extreme driving situations \cite{milliken:1995:RCVD}, past research on modeling of human drivers did not lead to a clear result.
	\change{Hence, dynamic vehicle simulations often use conventional controllers or reference maneuvers for standard driving situations with limited dynamics {\cite{ruf:2018:GTTV}}.
	As these vehicle controllers incorporate human behavior only to a certain extent, which is especially important in extreme driving situations, additional simulations with a human driver in the loop (HDIL) are required in the current development process.
	In particular, HDIL is mandatory for realistic simulations when testing driver assistance systems or when considering driving at the limits of friction {\cite{fritzsche:2017:SBDP}}.	
	The availability of a realistic and robust model of the human driver could extend and partially replace current HDIL evaluations.
	This facilitates an enhanced data-driven vehicle optimization process, combining a large number of simulations, reduced time effort, and a better consideration of human characteristics.}
	\par
	In this paper, we present a novel approach for driver behavior modeling in the context of car racing and evaluate it on a simplified vehicle model from \textit{OpenAI Gym} \cite{klimov:2016:CRVE}.
	\change{Our work is intended to pave the way for imitating human race car driver behavior in more complex settings like the Porsche Motorsport Driving Simulator environment {\cite{fritzsche:2017:SBDP}}, and to gain a better understanding of human race drivers in general.
	The flexibility of our approach potentially facilitates application to a number of other imitation learning tasks as well.
	In this context, future research could adapt the proposed framework to other problems, especially to engineering tasks involving human in the loop simulations.}
	\subsection{Problem Statement \& Notation}
		Driving a vehicle at the limits of handling, as present in car racing, is a complex skill which is influenced by numerous factors \cite{bentley:2011:USSC}. 
		The human driver's choice of actions does not solely depend on the current vehicle state, but also on prior knowledge and experience \cite{macadam:2003:UMHD}.	
		Learning such human driver behavior in the context of car racing is a challenging and complex problem, mainly due to four reasons:
		\begin{enumerate}
			\item multiple, partially unknown influencing factors exist \cite{bentley:2011:USSC},
			\item driver behavior is stochastic due to self-adaptation and human inconsistency \cite{bentley:2011:USSC}, \cite{bentley:1998:SSPR}, 
			\item even though two drivers achieve similar or equal performances, there could be considerable differences in their driving styles \cite{trzesniowski:2017:HRDA},
			\item small differences in the choice of actions potentially results in large path-deviations or even destabilization of the vehicle \cite{krumm:2015:DEAS}.
		\end{enumerate}
		A suitable driver model should be able to mimic the complex individual driver behavior and is required to be sufficiently robust at the limits of handling while being data efficient.
		Furthermore, it needs to capture the correlations and variance inherent in the demonstrations, in order to express human adaptation and imprecision.
		\par		
		Hence, we aim to learn a policy $\pi^M: \boldsymbol{s} \mapsto \boldsymbol{a}$ which maps the current state $\boldsymbol{s}$ to car control inputs $\boldsymbol{a}$, imitating an unknown stochastic expert policy $\pi^E$.
		 The policy $\pi^M$ is learned from demonstration data $\boldsymbol{\mathcal{D}}$ for each driver individually, where $\boldsymbol{\mathcal{D}}_{i,j}^k \subset \boldsymbol{\mathcal{D}}$ is the $i^{th}$ of $N_{\mathrm{laps}}$ demonstration laps on race track $j$ by driver $k$.
		It contains trajectories $\boldsymbol{\tau}_s$ of all available simulation states $\boldsymbol{s} \in \boldsymbol{\mathcal{S}}$.
		Finally, $\pi^M$ should be robust, generalizable and similar to the human driving style $\pi^E$, measured by a set of metrics $\boldsymbol{\mathcal{M}}$.
		The main contribution of this paper is hereby the development of a structured way of learning such a stochastic policy $\pi^M$ at the limits of handling.
	\subsection{Related Work}
		Human driver behavior and its modeling has been studied for decades, resulting in numerous approaches for specific use cases with different levels of abstraction.
		Already in 1985, Michon stated that \enquote{\textcolor{grey3P}{There appears to be a lack of new ideas in driver behavior modeling}} \cite{michon:1985:CVDB}.
		Extensive analyses of the driver, its physical limitations and attributes contributed to a better unterstanding of the complex human decision making process \cite{macadam:2003:UMHD}.
		Many approaches aim to manually construct and tune a mathematical formulation to imitate human driver behavior.
		In 1990, Hess and Modjtahedzadeh created a control theoretic model of driver steering behavior.
		The developed control scheme accounted for human limitations by tuning its frequency characteristics and produced human-like steering responses in a simulated lane-keeping driving task \cite{hess:1990:CTMD}.
		More advanced control algorithms like \textit{Model Predictive Control} (\textit{MPC}) allow to construct a more robust policy to imitate human driver behavior.
		Utilization of MPC for tracking a trajectory generated by a human driver in a simulator setup \cite{wei:2013:MHDB} or for tracking a reference path at the limits of handling \cite{novi:2019:CSFA} achieved quite promising results.
		\change{A virtual test driver model, developed by K\"{o}nig, uses an exact linearization controller and is intended to support the vehicle development process.
		This model was shown to work even at the limits of handling, but did not explicitly imitate human behavior {\cite{koenig:2009:VTQG}}.}
		\par
		\change{In the recent past, a number of machine learning approaches emerged which aim to construct a human-like policy directly from demonstration data for a simulated environment using imitation learning (IL).}
		Supervised methods could be utilized to learn higher-level control inputs which are subsequently mapped to car control inputs by conventional lower-level control policies \cite{cardamone:2009:LDTI}.
		Alternatively, a direct mapping from the current vehicle state to car control inputs can be learned with random forests \cite{cichosz:2014:ILCD}.
		A feedforward neural network is capable of tracking a target trajectory from a human demonstrator in a natural driving style \cite{wei:2013:MHDB}.
		Synthesized data combined with additional losses in an IL framework penalize undesirable situations, resulting in a more robust autonomous driving policy \cite{bansal:2018:CLDI}.
		\change{Apart from that, the \textit{Dataset Aggregation} (\textsc{DAgger}) method is a no-regret algorithm for online IL which guarantees to find a deterministic policy with good performance {\cite{ross:2011:RILS}}.
		Its extension \textsc{SafeDAgger} utilizes a safety classifier to determine if it is safe to use the imitating policy in the current situation and was tested in a car racing environment to imitate a rule-based driving controller {\cite{zhang:2017:QEIL}}.
		Furthermore, the \textit{Generative Adversarial Imitation Learning} (\textit{GAIL}) framework developed by Ho and Ermon {\cite{ho:2016:GAIL}} was tested for highway driving simulations {\cite{kuefler:2017:IDBG}} and learning specific driving styles {\cite{kuefler:2018:BIDM}}.
		In 2015, Kuderer et al. showed that feature-based inverse reinforcement learning is applicable to learn driving styles from demonstrations for autonomous vehicles {\cite{kuderer:2015:LDSA}}}.
		\par
		\change{These studies achieved impressive results in their specific applications.
		However, as none of this work explicitly encodes the variance of a human driver and is robust enough to drive in a car racing setting at the same time, they are of only limited applicability to the problem we are trying to solve.
		Hence, they do not fulfill the previously defined requirements completely, which motivates the search for other IL techniques originating from different fields of research.
		\par
		An interesting approach to represent human variability is found with \textit{Probabilistic Movement Primitives} (\textit{ProMPs}), which extend the widely used concept of movement primitives in robotics with a probabilistic trajectory description.}
		This framework directly encodes the mean and variance of human demonstrations and facilitates sequencing or blending of multiple, learned primitives \cite{paraschos:2013:PMP}, \cite{paraschos:2018:UPMP}.
		\par
		\change{Our work is intended to bridge the gap between a robust policy, which is working well at the limits of handling, and an accurate imitation of the human driver including its variability characteristics.
		}		
	\section{Probabilistic Modeling of Driver Behavior}
\begin{figure}[tb]
		\includegraphics[width=\columnwidth]{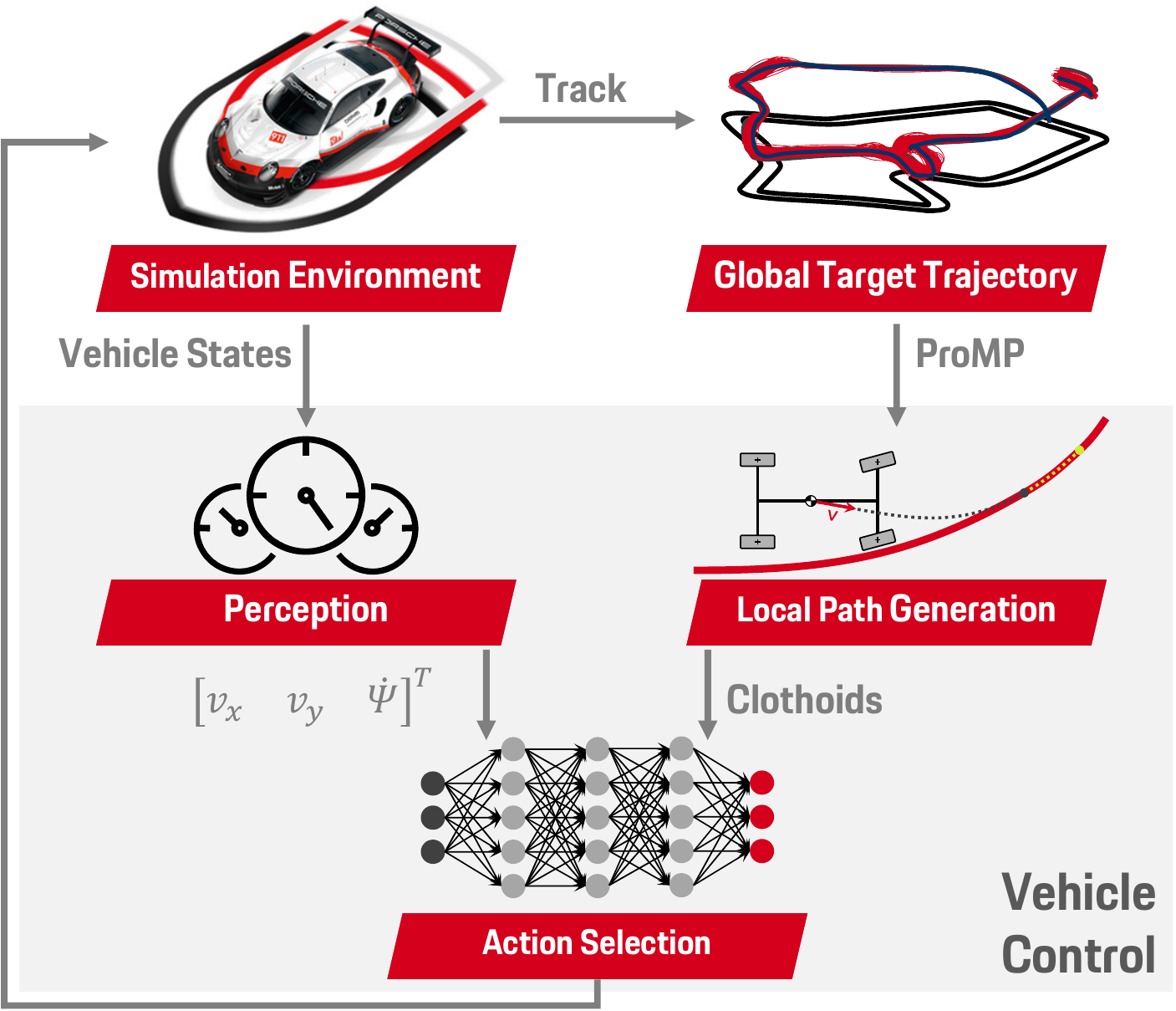}
		\caption{We introduce the novel ProMoD Architecture: for a specific track, ProMPs are utilized to describe a global target trajectory distribution in order to express human variability. Short-term \textit{Local Path Generation} constructs clothoids to represent a path from the current vehicle position to a future target position in a compact way.  A neural network maps this path information and basic vehicle states to car control actions, imitating the expert driving style. Picture for \textit{Simulation Environment} from \cite{porsche:2019:VMMD}.}
		\label{fig:ProMoD_architecture}		
	\end{figure}

	Every human driver exhibits a certain amount of variability, even when driving in a deterministic simulator environment.	
	This variability results from either intentional adaptation in order to optimize driving, or inconsistency which leads to slightly different control inputs \cite{bentley:2011:USSC}, \cite{bentley:1998:SSPR}.	
	Therefore, the human driver can be interpreted as a non-deterministic system which motivates a probabilistic modeling approach.	
	In this paper, we propose ProMoD as a modular framework to approach the problem of driver behavior modeling.
	\change{Inspired by autonomous driving system architectures {\cite{maurer:2016:ADTL}}, {\cite{nolte:2017:MPCB}}, {\cite{gonzalez:2015:RMPT}} and knowledge of driver behavior {\cite{bentley:2011:USSC}}, {\cite{bentley:1998:SSPR}}, {\cite{krumm:2015:DEAS}}, {\cite{kegelman:2018:LPRC}}, the basic model structure contains different modules as illustrated in Fig. {\ref{fig:ProMoD_architecture}}.}
	\par
	As we are operating in a simulated environment, all vehicle states and the position relative to the track are known exactly.
	\mbox{ProMoD} splits the task of driving into learning a target trajectory (\textit{Global Target Trajectory}) and following this trajectory by generating driving control inputs from a neural network (\textit{Vehicle Control}).
	ProMoD extends the basic idea of trajectory tracking \cite{wei:2013:MHDB} through a short-term \textit{Local Path Generation} and by a probabilistic representation of the target trajectory.
	This architecture enables the model to express self-adaptation and unintentional inconsistency, as discussed in the following.
	\subsection{Global Target Trajectory}
		Every driver has a quite precise idea where the car should go, not only in the current situation, but also some time ahead.
		Race drivers keep such a mental image in their heads for the whole track layout, knowing precisely where to brake, shift, turn in or accelerate at every point of the track \cite{bentley:2011:USSC}.
		However, these trajectories are driver specific, are modified over time due to adaptation, and the resulting movement of the car is strongly influenced by human inconsistency \cite{bentley:2011:USSC}, \cite{bentley:1998:SSPR}, \cite{trzesniowski:2017:HRDA}, \cite{krumm:2015:DEAS}.
		Hence, a universal single target trajectory does not exist, which motivates to learn a distribution of all demonstrated trajectories from a specific driver and track.
		ProMPs \cite{paraschos:2013:PMP} provide a suitable probabilistic representation of these demonstrated trajectories including motion variability and inherent correlations.
		\par
		In ProMoD, we \change{choose} to learn a single ProMP from demonstrations in order to represent the global target trajectory in a probabilistic way.
		This procedure is described in Algorithm \ref{alg:global_trajectory_planning}.	
		For a specific driver $k$ on track $j$ the process iterates over every single demonstration ${i \in \left\lbrace 1,~2,~ \hdots,~ N_{\mathrm{laps}} \right\rbrace}$ and performs a temporal modulation of the corresponding data $\boldsymbol{\mathcal{D}}_{i,j}^k$ initially.
		All trajectories are scaled into a normalized time frame with phase variable $z_i = \dot{z}_i ~ t ~ \in [0, 1]$, and phase change $\dot{z}_i = dz_i / dt = 1 / t_i^{\mathrm{end}}$,	where $t_i^{\mathrm{end}}$ denotes the length of demonstration $i$.		
		\begin{figure}[t]
		    \begin{minipage}{\columnwidth}
				\begin{algorithm}[H]
					\caption{Fit ProMP}
					\label{alg:global_trajectory_planning}
					\begin{algorithmic}
						\For{$i \leftarrow 1, N_{\mathrm{laps}}$}
							\State $\boldsymbol{\mathcal{T}}_i \leftarrow$ \Call{TemporalModulation}{$\boldsymbol{\mathcal{D}}_{i,j}^k$}
							\State $\boldsymbol{w}_i \leftarrow$ \Call{RidgeRegression}{$\boldsymbol{\mathcal{T}}_i$}
						\EndFor
						\State $\boldsymbol{\mu_w}, \boldsymbol{\Sigma_w} \leftarrow$ \Call{FitGaussian}{$\boldsymbol{w}_1, ... , \boldsymbol{w}_{N_{\mathrm{laps}}}$}
					\end{algorithmic}
				\end{algorithm}	
	    	\end{minipage}
  		\end{figure}		
		Then, this data is sampled equidistantly to $N_{\mathrm{samples}}$ grid points via cubic interpolation.					
		The resulting time normalized trajectories $\boldsymbol{\tau}_s$ of positions $x$ and $y$, and velocities $\dot{x}$ and $\dot{y}$ of the car are grouped into the target matrix
		${
			\boldsymbol{\mathcal{T}}_i =
			\begin{bmatrix}
				\boldsymbol{\tau}_x&
				\boldsymbol{\tau}_{\dot{x}}&
				\boldsymbol{\tau}_y&
				\boldsymbol{\tau}_{\dot{y}}
			\end{bmatrix}.
		}$
		\par	
		Subsequently, these time series are projected into a weight space via ridge regression on radial basis functions as detailed in \cite{paraschos:2013:PMP}.
		The weight vector $\boldsymbol{w}_i$ for the considered demonstration $i$ is obtained through ridge regression
		\begin{equation}
				\boldsymbol{w}_i = \left( \boldsymbol{\Phi} \boldsymbol{\Phi}^\intercal + \epsilon \boldsymbol{I} \right)^{-1} \boldsymbol{\Phi} \boldsymbol{\mathcal{T}}_i,
		\end{equation}		
		with regularization factor $\epsilon$ and the basis function matrix $\boldsymbol{\Phi}$.
		The basis function matrix contains the values of 38 radial basis functions at the predefined grid points in the phase space.
		Finally, a Gaussian distribution ${\boldsymbol{w} \sim \mathcal{N}(\boldsymbol{\mu_w},\,\boldsymbol{\Sigma_w})}$ is fitted over all $\boldsymbol{w}_i$ to calculate mean and covariance 
		\begin{align}
				\boldsymbol{\mu_w} &= \frac{1}{N_{\mathrm{laps}}}\sum_{i=1}^{N_{\mathrm{laps}}} \boldsymbol{w}_i,\\
				\boldsymbol{\Sigma_w} &= \frac{1}{N_{\mathrm{laps}}}\sum_{i=1}^{N_{\mathrm{laps}}} \left( \boldsymbol{w}_i - \boldsymbol{\mu_w} \right) \left( \boldsymbol{w}_i - \boldsymbol{\mu_w} \right)^\intercal
		\end{align}				
		of the weight vector $\boldsymbol{w}$.
		Both parameters directly encode the distribution of all demonstrated trajectories in the lower dimensional weight space.
		By drawing a sample weight vector $\boldsymbol{w}^*$ from $\mathcal{N}(\boldsymbol{\mu_w},\,\boldsymbol{\Sigma_w})$, we are able to reconstruct sample trajectories $\boldsymbol{\tau}_x^*,~ \boldsymbol{\tau}_{\dot{x}}^*,~ \boldsymbol{\tau}_y^*,~ \boldsymbol{\tau}_{\dot{y}}^*$ similar to the demonstrations through $\mathcal{T}^* = \boldsymbol{\Phi}^\intercal \boldsymbol{w}^*$.
		The execution speed, and hence the velocities $\dot{x}$ and $\dot{y}$ can be adjusted by adaptation of the phase change $\dot{z}$.
		Subsequently, this sample can be used as a global target trajectory for a specific driver on a predefined track which should be followed by the \textit{Vehicle Control}.
	\subsection{Vehicle Control}
		As the decision making processes of human drivers are complex and hard to model, we decide to utilize a neural network for the lower-level \textit{Action Selection}.
		Hence, the definition of a proper input representation is crucial.
		While the \textit{Perception}\footnotemark module represents the basic feedback from the car and gives information on the stability and current movement speed with the input vector $\boldsymbol{x}_P$, the \textit{Local Path Generation} is responsible for short term movement planning based on the global target trajectory and current vehicle position, and returns the input vector $\boldsymbol{x}_{LP}$.
		\change{A multilayer recurrent neural network is trained to map this information to the corresponding control actions, as detailed in the following.}
		\footnotetext{Note, that the meaning of perception differs from autonomous driving architectures. While we utilize the term for the basic feedback the driver receives from the car, autonomous driving architectures normally use it to denote the process of identifying obstacles, environmental conditions, and other road users \cite{maurer:2016:ADTL}.}
		\paragraph{Perception}
			The utilized perception feature vector
			${
				\boldsymbol{x}_P =
				\begin{bmatrix}
					v_x&
					v_y&
					\dot{\Psi}
				\end{bmatrix}^\intercal
			}$
			consists of the longitudinal and lateral velocities of the car $v_x$ and $v_y$, and its yaw rate $\dot{\Psi}$ in the vehicle-fixed coordinate system.
			These three basic vehicle states contain information on the vehicle stability and its future position and angle, and are therefore of great importance for the driver.			
			The current stability of the car is indicated by the ratio $ v_y / {v_x}$ and the yaw rate $\dot{\Psi}$, while the future position and yaw angle of the car can be predicted by integration of $v_x$, $v_y$ and $\dot{\Psi}$.
		\paragraph{Local Path Generation}
			\begin{figure}[tb]
			    \centering
			    \resizebox{\columnwidth}{!}{
				    \begin{turn}{-90}
					    \includegraphics{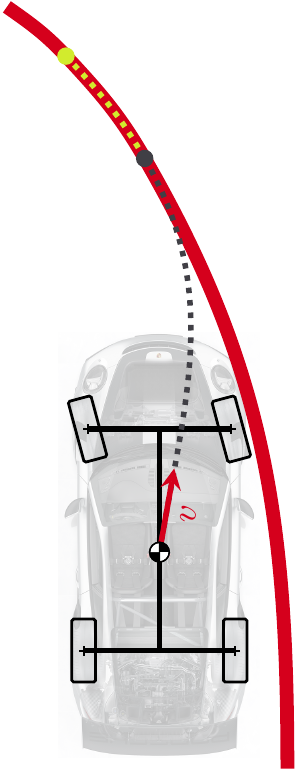}
	  				\end{turn}
  				}
			    \caption{\textit{Local Path Generation} using clothoids (own illustration with background picture adapted from \cite{porsche:2017:GT2R}): two clothoids are fitted to represent the short-term planning of the driver. The dark grey dotted line starts in the center of gravity of the car and ends on the global target trajectory shown in red after a short preview time. The second clothoid drawn in green extends this preview and gives information on the upcoming curvature of the global target trajectory.}
			    \label{fig:motion_planning}   
			\end{figure}
			The previously introduced \textit{Global Target Trajectory} finds a target driving line around the track which is similar to the demonstrated driving lines and represents the long-term path planning of the driver.
			During driving, a human mainly focuses at specific points of the upcoming track segment \cite{vanLeeuwen:2017:DRNR}, and the vehicle might deviate from the global path due to minor mistakes.
			Hence, the driver needs to plan a future driving line to the upcoming parts of the global trajectory, which motivates the development of a flexible and universal short-term path planning module.
			From a race driver's perspective, humans are using specific visual anchor points around the track and plan the short-term vehicle path relative to these points \cite{krumm:2015:DEAS}.
			\par
\begin{figure}[tb]
			\includegraphics[width=\columnwidth]{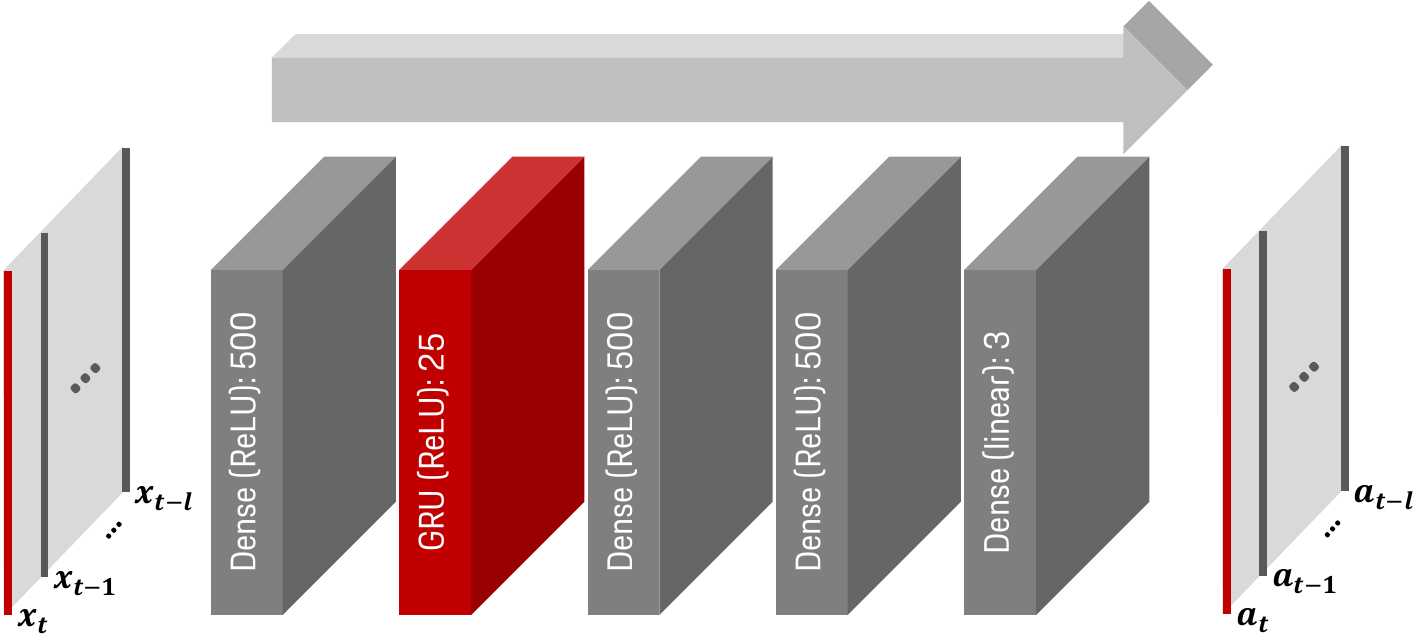}
			\caption{\change{We utilize a multilayer recurrent neural network to model the complex driver action selection process.
			Grey and red boxes show information on layer type, the utilized activation function, and the number of neurons.
			The Gated Recurrent Unit (GRU) layer incorporates short-term dependencies across multiple time steps, and is required to achieve a more realistic imitation of the human action selection process.
			For training, each sample contains features of the current and the previous $l=4$ time steps.{\fnref{note2}}
			}}
			\label{fig:nn_architecture}
		\end{figure}

			
			We base our approach on this assumption and utilize clothoids to provide a compact representation of the planned path.					
			A single clothoid is defined by three parameters, where $\kappa$ represents the curvature at the starting point, $\kappa'$ the change in curvature per unit of length, and $L$ the total length of this trajectory representation \cite{bertolazzi:2013:FAGF}. 
			The \textit{Local Path Generation} part of ProMoD consists of two clothoids as visualized in Fig. \ref{fig:motion_planning}.
			The first clothoid shown in dark grey is fitted to start in the center of gravity parallel to the velocity vector $v$ of the vehicle, and ends on the global target trajectory with a short preview time $P_1 = 250ms$ to provide essential information in order to neutralize lateral path deviation.
			Important information for the longitudinal control of the car is generated by the second clothoid shown in green, extending the preview of the upcoming target trajectory with $P_2 = 500ms$.\footnotemark
			\footnotetext{\label{note2}\change{$P_1$, $P_2$, and $l$ are treated as tunable hyperparameters. Optimal values might vary, depending on the simulator, vehicle, and human driver.}}
			The resulting \textit{Local Path Generation} features are defined as
			$
			{
			\boldsymbol{x}_{LP} =
				\begin{bmatrix}
					\kappa_1&
					\kappa'_1&
					L_1&
					\kappa_2&
					\kappa'_2&
					L_2&				
				\end{bmatrix}^\intercal
			},~
			$
			with subscripts $1$ and $2$ denoting the first and second clothoid respectively.
	\paragraph{Action Selection}
		While the previously presented modules provide a mapping from environmental states $\boldsymbol{s}$ and the ProMP, to relevant features $\boldsymbol{x}_{P}$ and $\boldsymbol{x}_{LP}$, the \textit{Action Selection} in ProMoD is intended to imitate the complex control selection process of the human driver.
		\change{We utilize a recurrent neural network to model the nonlinear mapping from the joint feature vector {$\boldsymbol{x}$} to the control inputs {$\boldsymbol{a}$}:}
		$
			\boldsymbol{x} =
			\begin{bmatrix}
				\boldsymbol{x}_{P}^\intercal&
				\boldsymbol{x}_{LP}^\intercal			
			\end{bmatrix}^\intercal
			\mapsto
			\boldsymbol{a} =
			\begin{bmatrix}
				\delta&
				g&
				b
			\end{bmatrix}^\intercal,
		$
		with steering angle {$\delta$} in {$\mathrm{deg}$}, and {$g \in \left[0,~1\right]$} and {$b \in \left[0,~1\right]$} denoting the gas and brake pedal actuation respectively.
		\change{As the decisions from a human driver do not solely depend on the current situation but also on previous sensory inputs and gathered experience, utilization of a recurrent architecture is important.
		The internal states of a Gated Recurrent Unit (GRU) layer might be interpreted to be related to the memory of a human driver.
		An overview of the neural network architecture is given in Fig. {\ref{fig:nn_architecture}}}
	\subsection{Training Procedure}
		\begin{figure}[t]
		    \begin{minipage}{\columnwidth}
				\begin{algorithm}[H]
					\caption{Train ProMoD}
					\label{alg:train_driver_model}
					\begin{algorithmic}
						\State $\boldsymbol{\mathcal{X}}, \boldsymbol{\mathcal{A}} \leftarrow \emptyset$
						\For{$j \leftarrow 1, N_{\mathrm{tracks}}$}
							\State $\boldsymbol{\mu_w}_{,j}, \boldsymbol{\Sigma_w}_{,j} \leftarrow$ \Call{Fit\_ProMP}{$j, k$}				
							\For{$i \leftarrow 1, N_{\mathrm{laps}}$}
								\State $\boldsymbol{X}_{P} \leftarrow$ \Call{Perception}{$\boldsymbol{\mathcal{D}}_{i,j}^k$}
								\State $\boldsymbol{X}_{LP} \leftarrow$ \Call{LocalPath}{$\boldsymbol{\mathcal{D}}_{i,j}^k, P_1, P_2$}
								\State $\boldsymbol{\mathcal{X}} \leftarrow \boldsymbol{\mathcal{X}} \cup
									\begin{bmatrix}
										\boldsymbol{X}_{P}\\
										\boldsymbol{X}_{LP}				
									\end{bmatrix}$
								\State $\boldsymbol{A} \leftarrow$ \Call{SelectActions}{$\boldsymbol{\mathcal{D}}_{i,j}^k$}						
								\State $\boldsymbol{\mathcal{A}} \leftarrow \boldsymbol{\mathcal{A}} \cup \boldsymbol{A}$
							\EndFor
						\EndFor
						\State  $\boldsymbol{NN} \leftarrow$ \Call{TrainNN}{$\boldsymbol{\mathcal{X}}, \boldsymbol{\mathcal{A}}$}
						\State $\pi^M \leftarrow \boldsymbol{\mu_w}, \boldsymbol{\Sigma_w}, \boldsymbol{NN}$
					\end{algorithmic}
				\end{algorithm}	
			\end{minipage}
		\end{figure}
		The training procedure for the complete driver model is described in Algorithm \ref{alg:train_driver_model}.
		For a specific driver $k$, the process iterates over all tracks $j$ and learns a global target trajectory distribution parametrized by $\boldsymbol{\mu_w}_{,j}, \boldsymbol{\Sigma_w}_{,j}$ for each track individually as defined in Algorithm \ref{alg:global_trajectory_planning}.
		Furthermore, we consider all demo data to extract the features $\boldsymbol{x}$ and corresponding actions $\boldsymbol{a}$ for each time step.
		While the perception features $\boldsymbol{x}_{P}$ are a subset of the vehicle state $\boldsymbol{s}$, the \textit{Local Path Generation} features $\boldsymbol{x}_{LP}$ are generated by fitting the clothoids with preview times $P_1$ and $P_2$ to the actually driven future driving lines when training the driver model.
		We aggregate all features $\boldsymbol{\mathcal{X}}$ and corresponding actions $\boldsymbol{\mathcal{A}}$ for all demo data from the human driver, and utilize it to train a neural network for \textit{Vehicle Control}.
		The policy $\pi^M: \boldsymbol{s} \mapsto \boldsymbol{a}$ of the resulting driver model is parametrized by $\boldsymbol{\mu_w}, \boldsymbol{\Sigma_w}$ which are track dependent, and the parameters of the neural network, which are valid for all tracks.
		The probabilistic representation of the \textit{Global Target Trajectory} module makes $\pi^M$ a stochastic policy which fulfills all previously defined requirements.
		For utilization of $\pi^M$ to drive a race car autonomously in a simulated environment, a global target trajectory is sampled from the previously learned ProMP initially.
		At each time step of the simulation, the current vehicle position is matched to the closest position on this sample trajectory in order to generate suitable previews for the \textit{Local Path Generation} features $\boldsymbol{x}_{LP}$.
		The resulting joint feature vector $\boldsymbol{x}$ of the current time step is subsequently mapped to car control actions $\boldsymbol{a}$ by the neural network.
		This process iterates until the vehicle finishes the current lap or leaves the track.
		Then, a new target trajectory could be sampled to start a new lap.	
	\section{Evaluation}
	In order to evaluate the performance of the proposed ProMoD framework, we benchmark our approach in a simulated car racing setting against basic IL algorithms.
	The approaches are tested on data from human drivers and on synthetic data, both generated in a simulated environment.
	\change{Three metrics are defined to assess the similarity between the drivers and the corresponding driver models.
	In addition, the imitation quality of ProMoD is compared to both baseline approaches using an experiment inspired by the \textit{Imitation Game} {\cite{turing:1950:CMI}}.}
	Finally, the robustness of ProMoD is verified on a vehicle model with reduced grip on multiple tracks.
	\subsection{Experiment Design}		
		\begin{figure}[tb]
			\includegraphics[width=\columnwidth]{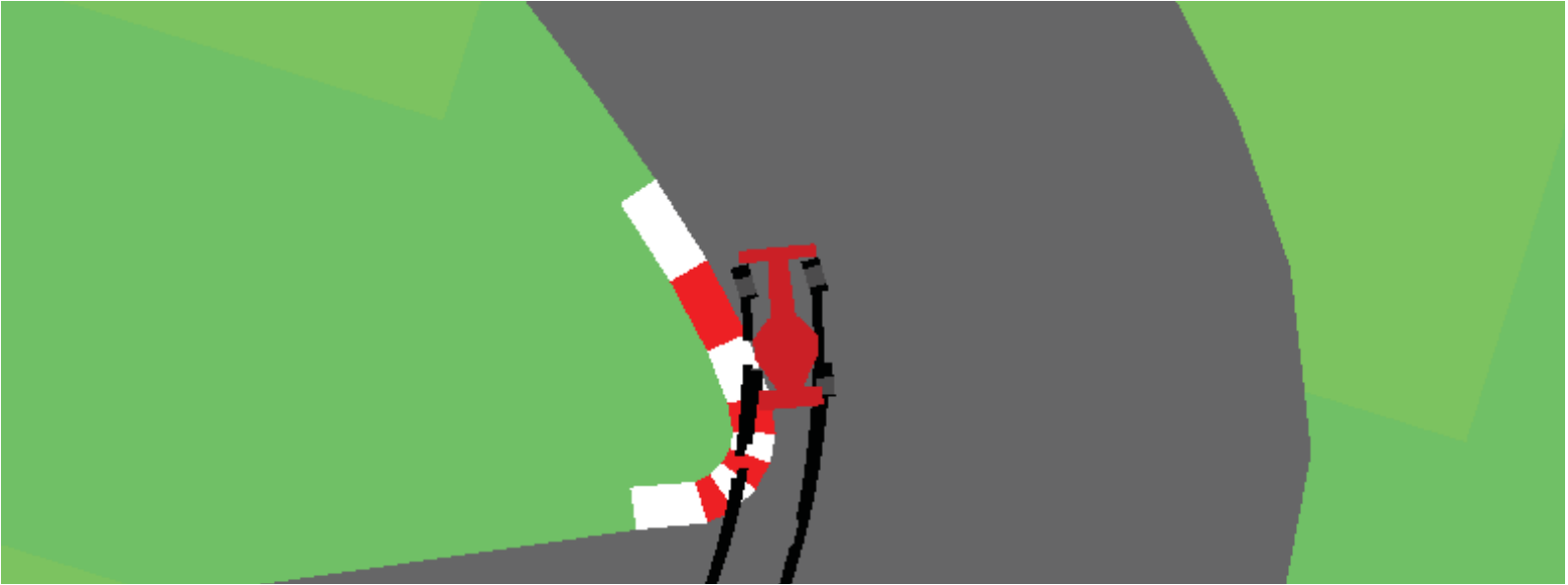}
			\caption{Simulation environment used for the experimental study \cite{klimov:2016:CRVE}. Black \change{tire} marks indicate that the vehicle is driven at the limits of handling.}
			\label{fig:toy_model}
		\end{figure}
		%
		\begin{figure*}[tb]
			\centering
			\includegraphics[width=0.8\textwidth]{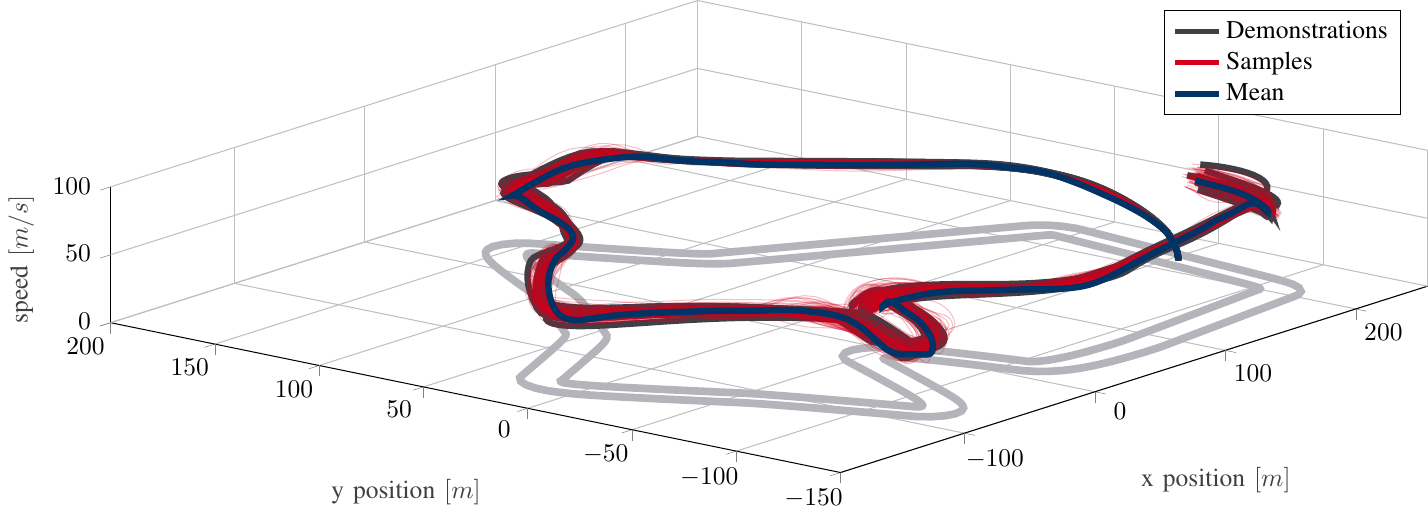} 
			\caption{3D visualization of \textit{Global Target Trajectory} generation using a ProMP: thick, light grey lines represent track boundaries, dark grey trajectories demo data from the human expert, and light red trajectories samples from the fitted ProMP. The mean trajectory is shown in blue.}
			\label{fig:driving_lines}
		\end{figure*}
		We utilize a slightly modified and reparametrized version of the \textit{OpenAI Gym CarRacing-v0 Environment} \cite{klimov:2016:CRVE} for the evaluation.\footnotemark
		\footnotetext{Modifications made to facilitate human control with continuous inputs. Changed parameters: FPS ${+200\%}$, road width ${+25\%}$, engine power ${-80\%}$, friction limit ${-30\%}$, brake force ${-87\%}$, $\textrm{steering ratio} = 8$}			
		This environment, shown in Fig. \ref{fig:toy_model}, simulates a rear-wheel driven race car and visualizes the current driving situation in a top-down view, enabling a human driver to control the car via a steering wheel and pedals.
		\change{The vehicle modeling approach has medium complexity, neglecting vertical dynamics but describing horizontal dynamics in a nonlinear way.
		Each tire is considered separately, and the resulting forces acting on the chassis are calculated based on the tire velocities which corresponds to a nonlinear two-track model {\cite{schramm:2010:MSDK}}.
		We choose this modeling approach as}
		\begin{enumerate*}
			\item \change{it is complex enough to express the fundamental vehicle characteristics when driving at the limits of handling, allowing over- and understeering of the race car,}
			\item \change{it is easily extensible and interpretable, and}
		 	\item \change{it is publicly available and could be used as a standard benchmark.}
		\end{enumerate*}
		\par
		\change{To evaluate the performance of ProMoD, we utilize data from four regular human drivers and synthetic data from a PID controller.}
		Each of these experts is requested to drive ten laps on five different tracks as fast as possible in the simulation environment.
		\change{Human drivers were given as much time as they requested in order to get used to the simulation environment and to each track.
		Besides that, they were allowed to restart a lap from the beginning at any time of this experiment.
		The PID controller was carefully tuned to drive fast and robust on a set of predefined driving lines.}
		\change{As will be shown in the next Subsection, those experts have considerably different driving styles and skill levels which should be imitated as closely as possible.}
		\par
		\change{We use this collected data to run an offline training of ProMoD, where Fig. {\ref{fig:driving_lines}} shows the learned ProMP for Driver 1 on Track 1.
		In addition, we test two basic IL algorithms as baseline approaches:}
		\begin{enumerate}
			\item \change{end-to-end supervised learning using the same set of training data as ProMoD,}
			\item \change{online learning using an adapted version of \textsc{DAgger} {\cite{ross:2011:RILS}}, which starts from the same data set, but gathers corrective demonstrations from the expert driver during the training procedure. These corrective actions should help to recover from mistakes when leaving the original data distribution.}
		\end{enumerate}
		\change{Both baselines utilize the same neural network architecture as given in Fig. {\ref{fig:nn_architecture}} and are trained on the same total amount of demonstration data as ProMoD for comparability reasons. 
		Due to the non-modular structure of both approaches, the feature definition differs from ProMoD.
		While utilizing identical features for the basic vehicle state $\boldsymbol{x}_{P}$, we replace the ProMoD path planning features $\boldsymbol{x}_{LP}$ with information on the vehicle position and the track layout according to knowledge on driver behavior:}
		\begin{enumerate*}
			\item \change{the lateral deviation from the track centerline,}
			\item \change{the heading error relative to the track centerline, and}
			\item \change{the curvature of the current and the upcoming 14 track segments, where the sign indicates corner directions.}
		\end{enumerate*}	
	\subsection{Similarity of Driving Styles}
\begin{figure}[tb]
			\centering
			\includegraphics[width=1.0\columnwidth]{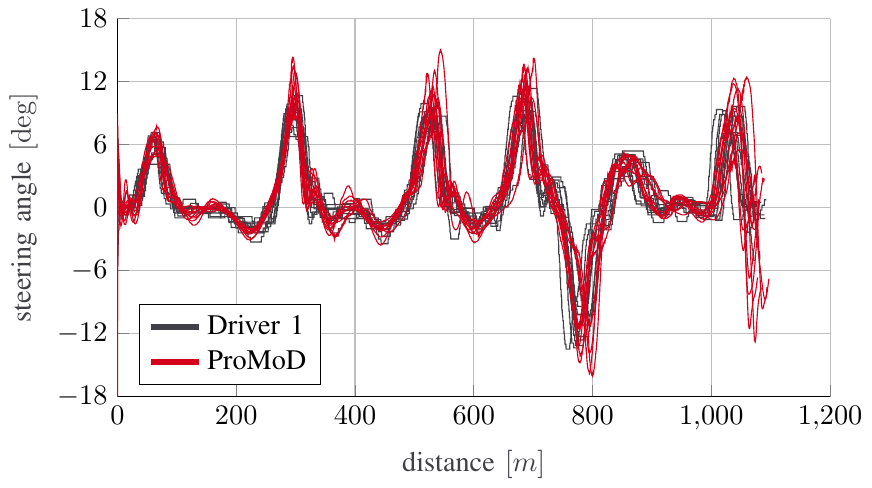}
			\caption{Comparison of the steering trajectories from human demonstrations, shown in dark grey, and generated by ProMoD, shown in red, on the reference vehicle. ProMoD is able to reproduce the distribution of the demonstrated steering trajectories with similar amplitudes and velocities. This indicates a similar style of steering the car.}
			\label{fig:actions_overlay}
		\end{figure}

		In order to judge the similarity between the learned policy $\pi^M$ and the expert policy $\pi^E$ in an objective way, we define a set of metrics $\boldsymbol{\mathcal{M}}$ based on the current state of research in driver analysis \cite{trzesniowski:2017:HRDA}.
		\change{These indicators quantify certain driver characteristics and allow objective comparisons of driving styles on a specific track.}
		\change{The laptime $t^{\mathrm{end}}$ can be considered as the most straightforward metric, where lower and less variant values indicate a better skilled and more experienced driver.}
		More complex expressions, depending on the demonstrated action trajectories are defined in the following.
		\begin{figure*}[!tb]
			\centering
			\includegraphics[width=0.86\textwidth]{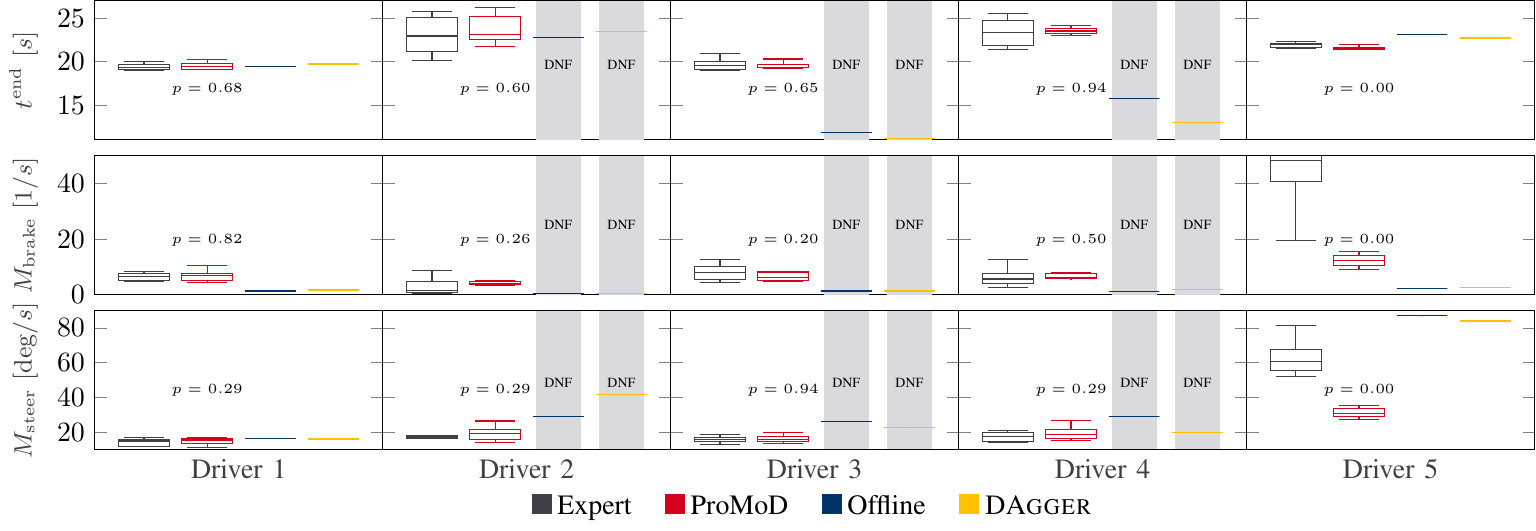} 
			\caption{Metrics evaluation for comparison of driving styles: \change{metrics of the four human drivers (Driver 1 to 4) and the synthetic driver (Driver 5) for ten laps on Track 1 are visualized with the box plots in dark grey, indicating different driving styles and skill levels.}
			Whisker ends represent the $\mathrm{min}$/$\mathrm{max}$ values. ProMoD is capable of successfully finishing ten laps for each driver with the resulting metric distribution shown in red.
			 We utilize a Kruskal–Wallis test (significance level of $\alpha=0.05$) to test the null hypothesis, that samples from the two groups (expert and ProMoD) are from populations with equal medians.
			 The resulting $p$-value for each driver-metric combination is printed next to the box plots.
			 For all human drivers we have $0.2 \leq p \leq 0.94$, indicating that there is no statistically significant difference in the median between the human drivers and ProMoD.
			 For Driver 5, however, there is very strong evidence to reject the null hypothesis.
			 This might result from the fact, that its synthetic control policy differs substantially from human behavior and can not be represented by ProMoD.
			 Direct supervised learning, visualized in blue, and \textsc{DAgger}, shown in yellow, did not finish (DNF) the laps for Driver 2, 3, and 4.
			 The lack of a defined target trajectory in the baseline approaches combined with less consistent demonstrations seems to reduce robustness.
			 The resulting metrics of the baseline aproaches are single values due to the deterministic policies and indicate a slightly reduced imitation accuracy.}
			\label{fig:metrics_comparison}
		\end{figure*}
		The steering aggressiveness
		\begin{align}
				M_{\mathrm{steer}} &= \frac{1}{\left\vert T_{\mathrm{int}} \right\vert} \int_{T_{\mathrm{int}}} \left\vert \dot{\delta} \right\vert dt ,\\
				T_{\mathrm{int}} &= \left\lbrace t \in \mathbb{R} ~ \Big\vert ~ 0 \leq t \leq t^{\mathrm{end}} \wedge \delta_{\mathrm{min}} < \left\vert \delta\left( t \right) \right\vert < \delta_{\mathrm{max}} \right\rbrace
		\end{align}
		integrates the absolute value of the steering velocity $\dot{\delta}$ during regular cornering defined by $\delta_{\mathrm{min}}$ and $\delta_{\mathrm{max}}$, and normalizes it with the total cornering duration $\left\vert T_{\mathrm{int}} \right\vert = \int_{T_{\mathrm{int}}} 1 ~ dt$.		
		Small values of $M_{\mathrm{steer}}$ correspond to a smoother style of steering the car, while higher values indicate a more aggressive steering or more corrections.
		The braking aggressiveness $M_{\mathrm{brake}}$ is defined equivalently with $\delta$ replaced by $b$.
		\change{Fig. {\ref{fig:metrics_comparison}} compares the driving styles of the five experts to the styles of the corresponding imitating driver models.}
		ProMoD is able to finish the track for all drivers, showing a slightly increased imitation accuracy.
		An overlay of steering trajectories from a human expert and the corresponding ProMoD model is shown in Fig. \ref{fig:actions_overlay}.
		The driver model is able to capture the steering amplitudes, velocities, and the corresponding distribution.	
		\par	
		\begin{table*}[tb]
			\centering
			\caption{\change{Turing-like test for race driver imitation: contingency tables for ProMoD and baseline approaches}}
			\renewcommand{\arraystretch}{1.15}
			\begin{tabular}{rr cc c}
			\toprule
				\multicolumn{2}{c}{\multirow{2}{*}{\textcolor{PorscheRed}{\textbf{ProMoD}}}}				& \multicolumn{3}{c}{\textbf{Truth}}												\\
																											&						& \textit{Human} 					& \textit{Robot} 					& $\Sigma$	\\ \midrule \midrule
				\multirow{3}{*}{\rot{\textbf{Judged}}}														&	\textit{Human}		& \textcolor{PorscheGreen}{$28\%$}	& \textcolor{PorscheRed}{$20\%$}	& $48\%$	\\ \cline{2-5}
																											&	\textit{Robot}		& \textcolor{PorscheRed}{$22\%$}	& \textcolor{PorscheGreen}{$30\%$}	& $52\%$	\\ \cline{2-5} \cline{2-5}
																											&	$\Sigma$			& $50\%$							& $50\%$ 										\\ \bottomrule
			\end{tabular}
			\hfill
			\begin{tabular}{rr cc c}
			\toprule
				\multicolumn{2}{c}{\multirow{2}{*}{\textcolor{blueP}{\textbf{Offline}}}}					& \multicolumn{3}{c}{\textbf{Truth}}												\\
																											&						& \textit{Human} 					& \textit{Robot} 					& $\Sigma$	\\ \midrule \midrule
				\multirow{3}{*}{\rot{\textbf{Judged}}}														&	\textit{Human}		& \textcolor{PorscheGreen}{$39\%$}	& \textcolor{PorscheRed}{$11\%$}	& $50\%$	\\ \cline{2-5}
																											&	\textit{Robot}		& \textcolor{PorscheRed}{$11\%$}	& \textcolor{PorscheGreen}{$39\%$}	& $50\%$	\\ \cline{2-5} \cline{2-5}
																											&	$\Sigma$			& $50\%$							& $50\%$ 										\\ \bottomrule
			\end{tabular}
			\hfill
			\begin{tabular}{rr cc c}
			\toprule
				\multicolumn{2}{c}{\multirow{2}{*}{\textcolor{PorscheYellow}{\textbf{\textsc{DAgger}}}}}	& \multicolumn{3}{c}{\textbf{Truth}}												\\
																											&						& \textit{Human} 					& \textit{Robot} 					& $\Sigma$	\\ \midrule \midrule
				\multirow{3}{*}{\rot{\textbf{Judged}}}														&	\textit{Human}		& \textcolor{PorscheGreen}{$38\%$}	& \textcolor{PorscheRed}{$12\%$}	& $50\%$	\\ \cline{2-5}
																											&	\textit{Robot}		& \textcolor{PorscheRed}{$12\%$}	& \textcolor{PorscheGreen}{$38\%$}	& $50\%$	\\ \cline{2-5} \cline{2-5}
																											&	$\Sigma$			& $50\%$							& $50\%$ 										\\ \bottomrule
			\end{tabular}		
			\parbox{\textwidth}{\footnotesize%
				\vspace{1em} 
				\change{
				We use a Turing-like test in order to judge, whether an imitating driver model can be distinguished from real drivers by human observers.
				This subjective analysis is done for ProMoD and both baseline approaches with identical conditions.
				For each driver imitation algorithm, we randomly draw ten demonstrations by human drivers and ten imitations generated by the corresponding approach from our database initially.
				These records are rendered such that the resulting videos show the car, the track, and the surroundings in a top-down view similar to Fig. {\ref{fig:toy_model}}.
				Subsequently, we shuffle each set of 20 videos to a random order and show those three sets of videos to five human experts (two drivers who are familiar with the toy model, and three engineers with knowledge of vehicle dynamics).
				Each of these experts is then asked to judge, which videos are generated by human drivers or by an imitating driver model (\textit{conditions}: no time limit, playback can be paused or restarted; \textit{available information}: each set contains ten videos from an unknown modeling approach and ten videos from human drivers with different driving styles and skills).
				The results are summarized in the three contingency tables above, comparing the assessment of the human observers to the ground truth, averaged over all experts.
				Green values in the upper left and lower right cells represent correctly classified human and robot demonstrations divided by the total number of demonstrations, while red values in the lower left and upper right corners indicate both types of misclassifications.
				Hence, low misclassification rates refer to a driver modeling algorithm which is easily distinguishable from human driving.
				Both baseline approaches yield total misclassification rates of $22\%$ and $24\%$ respectively, while ProMoD achieves $42\%$, indicating that our approach is considerably more difficult to distinguish from human driving.		
				}%
			}						
			\label{tab:turing_test}
		\end{table*}
		In addition to this objective assessment of similarity, we query the subjective perception of humans with a Turing-like test for race driver imitation.
		\change{For this purpose, we randomly select ten demonstrations from all human drivers and ten imitations from all five tracks for each approach.
		All videos are shown to five human experts which are asked to judge, whether a lap was generated by a human, or by a driver model.
		Table {\ref{tab:turing_test}} specifies the experiment protocol and lists the averaged results.
		In total, $40\%$ of the laps generated by ProMoD were considered to be driven by a human, and $44\%$ of the human driven laps were interpreted to be driven by the imitating ProMoD driver model.
		On the contrary, both baseline approaches achieved considerably lower, symmetric confusion rates of $22\%$ and $24\%$, respectively.
		This indicates, that the driving styles of ProMoD models are not easily distinguishable from real human driving in contrast to end-to-end supervised learning and \textsc{DAgger}.}
		\begin{figure}[tb]
		    \centering
			\includegraphics[width=1.0\columnwidth]{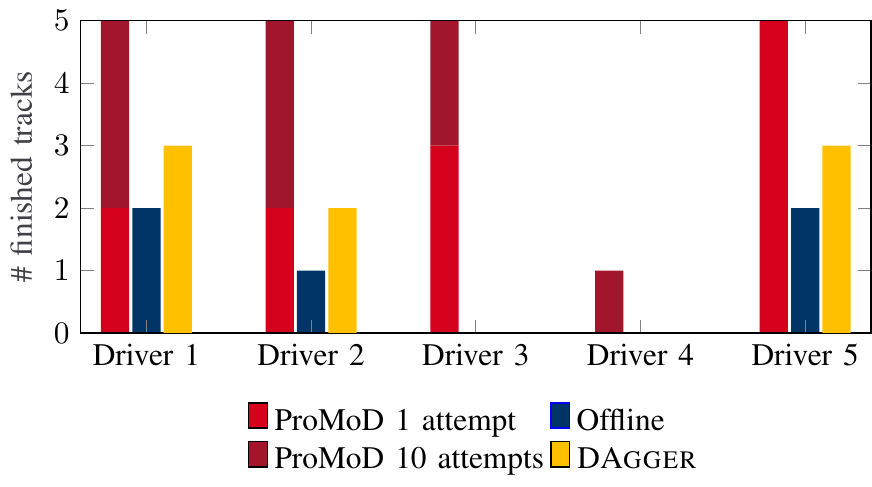}
			\caption{Robustness test: the three IL approaches are tested on five known tracks, but with the \change{tire} grip reduced to 90\%. This limitation makes the car more difficult to drive. Given 10 attempts (samples from \textit{Global Target Trajectory}), ProMoD is able to finish all tracks for all drivers except Driver 4, considerably outperforming the other IL algorithms.}
			\label{fig:robustness}
		\end{figure}		
	\subsection{Robustness}	
		In order to assess the robustness of ProMoD and the other IL approaches, we test them on the five known tracks, but with a reparametrized vehicle which is unknown to all driver models.
		We \change{choose} to reduce the \change{tire} grip to $90\%$, as this leads to an earlier saturation of the \change{tire} forces, resulting in a less stable vehicle which is considerably more difficult to drive.
		Hence, the number of successfully finished tracks with the modified vehicle is used as an indicator for robustness.
		Fig. \ref{fig:robustness} visualizes the results.
		As ProMoD is a probabilistic driver model, we evaluate the number of finished tracks with multiple samples from the learned global target trajectory distribution, here denoted as attempt.
		When considering only a single attempt, ProMoD is able to finish more tracks than the other IL approaches on average.
		For 10 attempts, ProMoD finishes all tracks for all drivers except Driver 4, indicating an increased robustness for varying conditions.
		Modification of other vehicle parameters in further tests yielded similar results.
	\section{Conclusion}
	In this paper, we present ProMoD as a modular framework for probabilistic modeling of driver behavior in a simulated environment.
	\change{Our approach is intended to mimic human driving styles and represent inherent variability, while being robust against varying vehicle parametrizations at the limits of handling.}
	To achieve this goal, we utilize ProMPs to learn the distribution of demonstrated driving trajectories, containing position and speed of the vehicle.
	Sampling from this distribution yields a target trajectory which is similar to the demonstrations.
	Information on the current vehicle state, as well as clothoids fitted to the upcoming target trajectory form the input representation for a neural network which is responsible for lower-level \textit{Action Selection}.
	Experiments in a simulated car racing setting compare the performances of ProMoD and basic IL algorithms.
	Our approach shows an increased imitation accuracy and robustness, potentially allowing to assess the drivability and performance of new vehicle parametrizations.
	\change{In a Turing-like test we analyze the visual distinguishability of laps generated by human drivers and imitating driver models.
	The experiments indicate that ProMoD models are considerably more difficult to distinguish from human \mbox{drivers compared to both baseline approaches.}} 
	\par
	\change{As first experiments in more complex environments already yield promising results as well, we expect ProMoD to be suited for higher dimensional problems, too.
	Its flexibility enables our approach to be applied to other imitation learning tasks.
	Especially simulation-based design optimization processes, for which an integration of human control of any kind of vehicle is essential, could benefit from ProMoD.
	It might be possible, for instance, to adapt ProMoD to a simulated drone or airplane flying scenario.
	This could be achieved by adding the vertical dimension to the global target trajectory and modifying the local path generation and perception features.
	Furthermore, parts of our framework might be used to extend existing autonomous driving architectures to yield a more human-like way of driving. 
	At the moment, however, ProMoD itself does not explicitly consider human self-optimization and is limited to single-driver scenarios on known tracks.
	Multi-driver scenarios are currently not supported, as our approach does not include human interactions which are hard to model in high dynamic environments.
	}
	\par
	\change{Nevertheless, due to the modular architecture of ProMoD, it is possible to extend our approach in various ways in future work.}
	The probabilistic representation of the target trajectory allows to iteratively adapt the driving line in order to represent human self-optimization.
	This could be achieved via conditioning of the learned ProMP or by situative modulation of the phase velocity $\dot{z}$.
	Furthermore, the global target trajectory distribution could be generated by a sequence of multiple ProMPs which represent basic driving maneuvers, resulting in a generalizable driver model for unknown tracks.
	\change{Utilization of dynamic time warping for temporal alignment of the demonstrations, a method for learning $P_1$, $P_2$, and $l$, as well as an extended input representation might be able to further enhance imitation accuracy.}
	Finally, ProMoD should be tested on the complex Porsche Motorsport Driving Simulator environment with data from professional human race car drivers, checking the applicability of our approach for driving at the limits of handling.	
	\bibliography{main}
\end{document}